\documentclass[11pt,a4paper]{article}
\usepackage[hyperref]{naaclhlt2018}
\usepackage{times}
\usepackage{latexsym}
\usepackage{filecontents}
\usepackage{pgfplots}
\pgfplotsset{compat=1.14}
\usepackage{verbatim}
\usepackage{float}
\usetikzlibrary{shapes.geometric,arrows, calc, backgrounds}
\usepackage{url}
\usepackage{multirow}
\usepackage{natbib}
\usepackage{silence}
\usepackage{etoolbox}
\apptocmd{\sloppy}{\hbadness 10000\relax}{}{}

\aclfinalcopy 

\title{Frisio41 at SemEval-2019 Task 6:  Combination of multiple Deep Learning architectures for Offensive Language Detection in Tweets}

\author{Batuhan Guler \\
  Imperial College London \\
  {\tt bbg2218@ic.ac.uk} \\\And
  Alexis Laignelet \\
  Imperial College London \\
  {\tt acl18@ic.ac.uk} \\\And
  Nicolo Frisiani \\
  Imperial College London \\
  {\tt nf1518@ic.ac.uk} \\
  }

\date{}

\begin{document}

\maketitle
\begin{abstract}
  This report contains the details regarding our submission to the OffensEval 2019 (SemEval 2019 - Task 6) \cite{offenseval}. The competition was based on the Offensive Language Identification Dataset \cite{OLID}. We first discuss the details of the classifier implemented and the type of input data used and pre-processing performed. We then move onto critically evaluating our performance. We have achieved a macro-average F1-score of 0.76, 0.68, 0.54, respectively for Task a, Task b, and Task c, which we believe reflects on the level of sophistication of the models implemented. Finally, we will be discussing the difficulties encountered and possible improvements for the future. 
\end{abstract}

\section{Introduction}
With the rise of online activity and the exploding popularity of social networks, offensive language has become an ever more serious hurdle, that raised concerns within many governments, companies, communities and organizations. For this reason, a group of researchers organized an Offensive Language Detection challenge \cite{offenseval} as part of the SemEval 2019 competition. This paper covers the details of our submission to this competition.  

\section{Classifier details}
\subsection{Initial experiments}
Throughout the course of this project we experimented with various different classifiers. Using scikit-learn\footnote{ \url{https://scikit-learn.org/}}, we started with a bag-of-words representation of tweets and a Random Forest classifier to set ourselves a baseline.

We then moved onto more sophisticated architectures using Keras with a Tensorflow backend, first implementing a Convolutional Neural Network (CNN) \cite{CNN}. Convnets have first transformed the field of image recognition with 2D Convolution, and have then been applied to sentiment classification in NLP with 1D Convolution. The CNN enabled us to capture the relationship between nearby words in a sentence. We then moved onto exploring Recurrent Neural Networks (RNN) as these allow the model to 'remember' previous inputs, which is key for understanding human languages. In fact, humans don't simply read a sentence word-by-word out of context, but understand the intrinsic meaning of each word based on the previous words in the sentence. This is the effect that RNNs attempt to recreate. In the end, we settled on a intricate combination of various architectures, including both a CNN and an RNN, as these seemed to give us the best performance and are a some of the most up-to-date techniques in NLP.

\subsection{Model in detail}
Our best-performing model is combination of a bi-directional RNN using LSTM cells, a CNN, and a Feed Forward Neural Network (FFNN). 
For the hyper-parameter search of our model we used Bayesian Optimization (BO), as this is a state-of-the-art technique in its field \cite{bayesian}. 
After 10 epochs of BO on a wide range of possible values, we landed on the following optimal parameters: learning rate = 0.001 and a weight decay of 6e-10. The weight decay obtained is really small and hence we replaced it with 0. Indeed, the purpose of weight decay is to prevent overfitting. However, we are already using Dropout technique (with a rate of 0.5) to prevent overfitting, which is probably what makes weight decay unnecessary.

\begin{figure}[H]
\centering
\tikzstyle{line} = [ draw, -latex']  
\tikzstyle{process} = [rectangle, fill=white, text centered, draw=black,thick]  
\tikzstyle{inter} = [rectangle, fill=white, text centered, draw=black,thick] 
\tikzstyle{term} = [rectangle, fill=white, text centered]

\begin{tikzpicture}[node distance=0.85 cm, auto]

  \node [term]           (puc)  {Output};  
  \node [process, below of=puc]  (wdt)  {FFNN};  
  \node [process, below of=wdt]  (port) {CNN};  
  \node [process, below of=port] (loop) {RNN};  
  \node [inter, below of=loop] (bon) {Embeddings};
  \node [term, below of=bon] (tweet) {Tweets};

  \path [line] (wdt)  -- (puc);  
  \path [line] (port)  -- (wdt);  
  \path [line] (loop) -- (port);  
  \path [line] (bon) -- (loop);  
  \path [line] (tweet) -- (bon);

\end{tikzpicture}
\caption{General architecture}
\label{fig:gen_archi}
\end{figure}
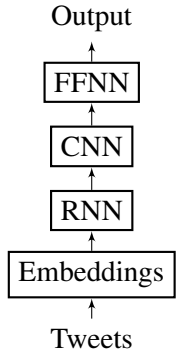

Figure \ref{fig:gen_archi} shows an outlook of the general architecture of our final model.

For the activation functions, we use the standard Tanh in the LSTM cells, Relu for all the hidden layers in the FFNN, and then softmax for the output layer. Furthermore, the word embeddings (covered in more detail later) are set to be trainable, so that the model will also update them while training for classification.

Table \ref{table:layers} shows the inner structure of the final network in more detail:
\begin{table}[H]
\begin{center}
\begin{tabular}{|l|c|c|}
\hline \bf Layer & \bf Output shape & \bf Params \\ \hline
embedding & (63, 100) & 2125100 \\
spatial\_dropout & (63, 100) & 0 \\
bidirectional &  (63, 256) & 234496 \\
conv & (62, 64)  & 32832 \\
max\_pooling&(64)& 0\\
average\_pooling&(64)&0 \\
concatenate&(128)&0\\
dense&(10)&1290\\
dense&(1)&11\\
\hline
\end{tabular}
\end{center}
\caption{Different layers detailed}
\label{table:layers}
\end{table}

The bidirectional RNN with 100 hidden units is followed by a convolution. Then, separately, we perform a maxpool and an average pooling, and concatenate the results. These are then fed into a 2 layer fully-connected network with 10 hidden neurons. 

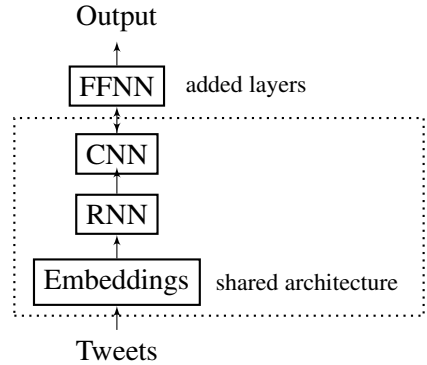
\begin{figure}[H]
\centering
\tikzstyle{line} = [ draw, -latex']  
\tikzstyle{process} = [rectangle, fill=white, text centered, draw=black,thick]  
\tikzstyle{inter} = [rectangle, fill=white, text centered, draw=black,thick] 
\tikzstyle{term} = [rectangle, fill=white, text centered]
\tikzstyle{term2} = [rectangle, text centered]

\begin{tikzpicture}[node distance=0.9cm, auto]

  \node [term]           (puc)  {Output};  
  \node [process, below of=puc]  (task)  {FFNN}; 
  \node [process, below of=task]  (wdt) {CNN};
  \node [process, below of=port] (loop) {RNN};  
  \node [inter, below of=loop] (bon) {Embeddings};
  \node [term, below of=bon] (tweet) {Tweets};

  \begin{scope}[node distance=2.5cm]
      \node [term2, right of=bon] (share) {\footnotesize shared architecture};
  \end{scope}
  
    \begin{scope}[node distance=1.7cm]
       \node [term2, right of=task] (add) {\footnotesize added layers};
  \end{scope}

    \begin{scope}[on background layer]
          \draw[black,thick,dotted] ($(wdt.north west)+(-0.8,0.2)$)  rectangle ($(share.south east)+(0.2,-0.2)$);
    \end{scope}

  \path [line] (task)  -- (puc); 
  \path [line] (wdt)  -- (task);  
  \path [line] (port)  -- (wdt);  
  \path [line] (loop) -- (port);  
  \path [line] (bon) -- (loop);  
  \path [line] (tweet) -- (bon);   
  block

\end{tikzpicture}
\caption{General architecture with shared layers for tasks A, B and C}
\label{fig:graph_archi}
\end{figure}

To improve the performance, we used a technique transfer learning \cite{transfer}, which consists of storing the knowledge gained while solving task A and applying it to the different but related tasks B and C. This consists in reusing the initial and middle layers of our model and adding a new feed forward neural network at the end, different for each task. 
Hence, our final architecture should be redrawn as in Figure \ref{fig:graph_archi}.

This technique is particularly effective when there are few labels for later tasks (which is the case in this problem, since the minority labels in tasks B and C appear as little as 420 times).

\section{Type of Input Data}
\subsection{Bag of Words}
Initially, we used a simple bag of words representation. This  representation uses only the co-occurrences of words and serves as a baseline. We first transform our corpus (tweets) into a document-term matrix, where each cell represents the number of occurrence of each word in the document. We can then simply fit a classifier on this matrix. We have chosen to use a random forest classifier, as it is a simple yet powerful model. Some transformations, like TF-IDF or feature selection using $\chi{2}$ test could have been used but we decided to have a simple model as a baseline.

\subsection{Embeddings}
As we then moved onto more complex architectures, we started using word embeddings \cite{word2vec}. Within the realm of word embeddings, we tried two different alternatives: we imported pre-trained embeddings from Stanford \cite{glove}, and we tried training our own embeddings from the given data. 

Our own training was performed on the whole dataset using a fast implementation for learning representations has been proposed with Word2Vec: FastText \cite{fasttext} from the library Gensim. This algorithm uses a fake task to learn the embeddings. This fake task is a one hidden layer neural network that tries to predict a word given its context (CBOW version of the algorithm). Each word is projected into a higher dimensional space in the hidden layer. FastText uses this algorithm, but instead of using only the word-context pairs, it uses the n-grams of characters of each of these word. This will enrich the information that we have for each word. What this means in practice is that if a tweet were to, for example, use the word 'newcar', the algorithm will not classify it as a complete new word but using the n-grams that composes this word, it will be able to understand that this word is composed by 'new' and 'car' and understand that it represents the concept of a new car. This is particularly well suited for the task at hand, because Tweets (especially offensive ones) tend to showcase a very creative use of the English language. We learnt our model with n-grams for n between 3 and 6.

As for the Stanford pre-trained embeddings (GloVe), of all the possible GloVe embeddings available, we used the GloVe.twitter embeddings. The choice comes from the fact that the data of our task is made purely of tweets, which have a systematically different use of the English language from Wikipedia or newspaper articles. Hence pre-trained embeddings can only be used if they were trained using tweets. 

However, despite our expectations, either because of the property of FastText described above, or maybe due to the fact that the tweets for this task use a mildly different English vocabulary from the one used to train the GloVe.twitter embeddings, the embeddings trained directly on our data seemed to perform surprisingly well, better than the GloVe ones. Hence we decided to settle on these for our final solution. 

In addition, we have seen that fine-tuned embeddings lead to better results than fixed embeddings. This was not a surprise, as task-specific embeddings are in general more performing.

\section{Pre-processing}
\subsection{Base-cleaning}
First, we removed the unlabeled tweets for each task.
Then we proceeded onto pre-processing  the data. The idea was to put all the tweet into lowercase, and delete useless punctuation, to make them as uniform and standardized as possible so that the model could compare them easily. This means we had to remove {\em '\#'} and {\em'@'}, remove repetitive {\em '@USER'}, and add space between words and punctuation (such as {\em '!'}  or {\em '?'}).

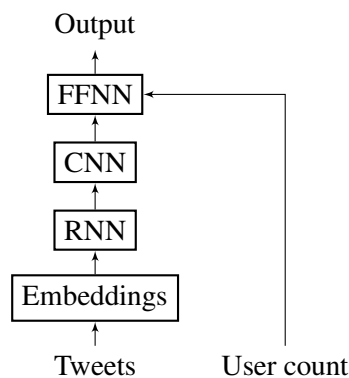
\begin{figure}[H]
\centering
\tikzstyle{line} = [ draw, -latex']  
\tikzstyle{process} = [rectangle, fill=white, text centered, draw=black,thick]  
\tikzstyle{inter} = [rectangle, fill=white, text centered, draw=black,thick] 
\tikzstyle{term} = [rectangle, fill=white, text centered]

\begin{tikzpicture}[node distance=0.9cm, auto]  
 
  \node [term]           (puc)  {Output};  
  \node [process, below of=puc]  (wdt)  {FFNN};  
  \node [process, below of=wdt]  (port) {CNN};  
  \node [process, below of=port] (loop) {RNN};  
  \node [inter, below of=loop] (bon) {Embeddings};
  \node [term, below of=bon] (tweet) {Tweets};
  
  \begin{scope}[node distance=2.5cm]
    \node [term, right of= tweet] (user) {User count};
  \end{scope}
 
  \path [line] (wdt)  -- (puc);  
  \path [line] (port)  -- (wdt);  
  \path [line] (loop) -- (port);  
  \path [line] (bon) -- (loop);  
  \path [line] (tweet) -- (bon);   
  \path [line] (user) |- (wdt); 
  
\end{tikzpicture}
\caption{General architecture taking into account the user count}
\label{fig:archi_user}
\end{figure}

Additionally, we tried creating an extra column to store the number of {\em '@USER'} for each tweet. The initial thought was that there could be several advantages of doing so: we thought it would be an important feature, which should be captured by the model, and we didn't want to lose this information but still wanted to remove redundant {\em '@USER'} during our cleaning process, as these were playing and overly emphasized role in the model. The architecture of the neural network was hence updated as shown in Figure \ref{fig:archi_user}.

However, when we implemented this architecture, it didn't seem to bring any improvement to the model. This might be because, while the average count of users seemed to be significantly different between classes, the standard deviation of this count is so high that the model didn't manage to extract any meaningful information from it (see Table \ref{table:user_count}).

\begin{table}[H]
\begin{center}
\begin{tabular}{|l|cc|}
\hline
\textbf{Label}            & \textbf{mean} & \textbf{std} \\ \hline
Task A - OFF          &   2.03            &       3.93       \\
Task A - NOT            &     2.75          &        6.26     \\
Task B - UNT            &  1.95             &      4.46        \\ 
Task B - TIN           &   2.04            &       3.85       \\ 
Task C - IND           &   1.90            &       4.45       \\ 
Task C - OTH            &  2.09             &      3.96        \\ 
Task C - GRP            &  2.33             &      4.58        \\ \hline
\end{tabular}
\end{center}
\caption{'user\_count' statistics per class per task}
\label{table:user_count}
\end{table}

Indeed, these statistics reveal that, whereas the difference seems significant for average user count between (for example) offensive and non-offensive tweets (task a), that difference amounts to less than 0.12 standard deviations of the distribution of user counts in non-offensive labels. 

To summarize, tweets, before cleaning, typically look like this:
{\em '@USER @USER @USER It should scare every American!  She is playing Hockey with a warped puck!'}
Instead, after applying our cleaning process, the resulting tweet will be:
{\em 'user it should scare every american ! she is playing hockey with a warped puck !'}

At an early stage of the project, we also tried the library Spacy to remove stop words from the corpus of tweets. However, it appears that doing so actually removed relevant content from the tweets, which resulted in a worsened performance.

\subsection{Imbalanced dataset}
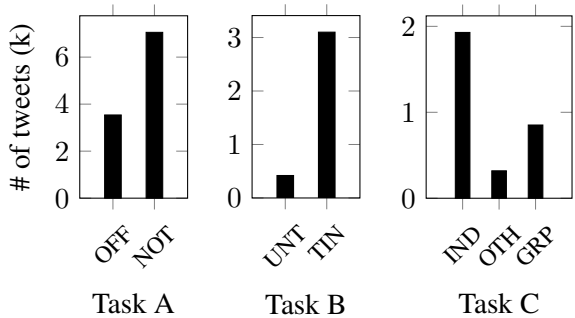
\begin{figure}[H]
\minipage{0.34\linewidth}
\centering

        \begin{tikzpicture}
        \begin{axis}[
            width=3cm,
            height=4cm,
            ybar,
            bar width=0.4,
            ymin=0,
            ytick={0,2,4,6,8},
            xticklabel style={rotate=45,font=\footnotesize},
            xticklabels from table={data.dat}{x}, 
            xtick=data, 
            enlarge x limits=0.8, 
            xlabel=Task A,
            ylabel=\# of tweets (k)
            ]
            \addplot[fill=black] table[x expr=\coordindex,y=y] {data.dat};
        \end{axis}
        \end{tikzpicture}

\endminipage\hfill
\minipage{0.3\linewidth}
\centering

        \begin{tikzpicture}
        \begin{axis}[
            width=3cm,
            height=4cm,
            ybar,
            bar width=0.4,
            ymin=0,
            ytick={0,1,2,3,4},
            xticklabel style={rotate=45,font=\footnotesize},
            xticklabels from table={data.dat}{x}, 
            xtick=data, 
            enlarge x limits=0.8, 
            xlabel=Task B,
            ]
            \addplot[fill=black] table[x expr=\coordindex,y=y] {data.dat};
        \end{axis}
        \end{tikzpicture}
\endminipage\hfill
\minipage{0.34\linewidth}
\centering

        \begin{tikzpicture}
        \begin{axis}[
            width=3.5cm,
            height=4cm,
            ybar,
            bar width=0.4,
            ymin=0,
            ytick={0,1,2,3,4},
            xticklabel style={rotate=45,font=\footnotesize},
            xticklabels from table={data.dat}{x}, 
            xtick=data, 
            enlarge x limits=0.5, 
            xlabel=Task C,
            ]
            \addplot[fill=black] table[x expr=\coordindex,y=y] {data.dat};
        \end{axis}
        \end{tikzpicture}
\endminipage
        \caption{Number of tweets per class per task}
        \label{fig:imbalanced}
\end{figure}

The distribution of the labels among the three datasets are as shown in Figure \ref{fig:imbalanced}.
As it can be seen, the datasets are heavily imbalanced. For task A, there are twice more non offensive examples (NOT) than offensives ones (OFF), as for task B, there are six times more tweets targeted to an individual (TIN) than non targeted (UNT) ones. For task C, classes are not equally represented either.

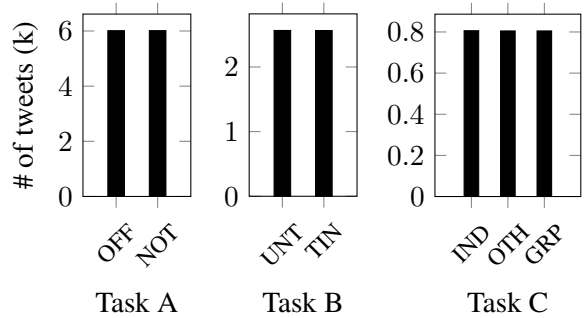
\begin{figure}[H]
\minipage{0.34\linewidth}
\centering

        \begin{tikzpicture}
        \begin{axis}[
            width=3cm,
            height=4cm,
            ybar,
            bar width=0.4,
            ymin=0,
            ytick={0,2,4,6,8},
            xticklabel style={rotate=45,font=\footnotesize},
            xticklabels from table={data.dat}{x}, 
            xtick=data, 
            enlarge x limits=0.8, 
            xlabel=Task A,
            ylabel=\# of tweets (k)
            ]
            \addplot[fill=black] table[x expr=\coordindex,y=y] {data.dat};
        \end{axis}
        \end{tikzpicture}
\endminipage\hfill
\minipage{0.26\linewidth}
\centering

        \begin{tikzpicture}
        \begin{axis}[
            width=3cm,
            height=4cm,
            ybar,
            bar width=0.4,
            ymin=0,
            ytick={0,1,2,3,4},
            xticklabel style={rotate=45,font=\footnotesize},
            xticklabels from table={data.dat}{x}, 
            xtick=data, 
            enlarge x limits=0.8, 
            xlabel=Task B,
            ]
            \addplot[fill=black] table[x expr=\coordindex,y=y] {data.dat};
        \end{axis}
        \end{tikzpicture}
\endminipage\hfill
\minipage{0.36\linewidth}
\centering

        \begin{tikzpicture}
        \begin{axis}[
            width=3.5cm,
            height=4cm,
            ybar,
            bar width=0.4,
            ymin=0,
            ytick={0,0.2,0.4,0.6,0.8},
            xticklabel style={rotate=45,font=\footnotesize},
            xticklabels from table={data.dat}{x}, 
            xtick=data, 
            enlarge x limits=0.5, 
            xlabel=Task C,
            ]
            \addplot[fill=black] table[x expr=\coordindex,y=y] {data.dat};
        \end{axis}
        \end{tikzpicture}
\endminipage
\caption{Number of tweets per class per task after up-sampling and down-sampling}
\label{fig:balanced}
\end{figure}

This can drastically affect the performance of the model. To prevent that, oversampling and undersampling were used to make the classes' weights equal. The function performing the over and under sampling takes an input representing the percentage of undersampling desired $p_u$ (1 - $p_u$ will then be the percentage of oversampling) to reach the equal weighting between classes.
This number represents the trade-off between deleting potentially relevant information from the majority class or producing a minority class with too many duplicate examples. Hence, to choose it, we ran various iterations of cross validation performed on the random forest model, which resulted in $p_u =0.3$ for task A, $p_u = 0.2$ for task B, and $p_u = 0.7$ for task C. See resulting balanced datasets in Figure \ref{fig:balanced}.

\section{Performance Evaluation}

In order to evaluate our model, we split out data into train and validation, with 20\% of the total dataset assigned to validation. The test samples are then provided by the platform.

Table \ref{table:results_methods} shows the performance of our initial models: the random forest on the bag of words and the CNN on the word embeddings
\begin{table}[H]
\begin{center}
\begin{tabular}{|l|cc|}
\hline \bf Task & \bf Random forest & \bf CNN \\ \hline
Task A & 0.69 & 0.71 \\
Task B & 0.56 & 0.54 \\
Task C & 0.50 & 0.49 \\
\hline
\end{tabular}
\end{center}
\caption{Macro-avg F1 performance on validation test over different tested models}
\label{table:results_methods}
\end{table}

While the above results seem quite surprising (indeed we would expect a CNN to outperform a Random Forest in sentence classification \cite{CNN}), the reason behind them is pretty simple: due to the high speed of computation of the random forest (a handful of seconds compared to almost 10 minutes for the other deep learning models we implemented), throughout the project it was used as the 'validation model' to tune the data processing and various shared hyper-parameter. This means that the way we prepare the data will be much more tailored (if not over-fitted) to the random forest than to the CNN, which makes the latter seem like it has a worse performance.

Finally, the performance of the final model across the three tasks is summarized in Tables \ref{table:final_validation} and \ref{table:final_test}. This performance was achieved by performing 2 epochs of training for Task A, 7 epochs for Task B and 3 epochs for Task C. The number of epochs were chosen using the early-stopping algorithm, which stops the training iterations when the accuracy on the validation dataset has reached its peak (when it starts to over-fit).

\begin{table}[H]
\begin{center}
\begin{tabular}{|l|ccc|}
\hline \bf Label & \bf precision & \bf recall & \bf F1\\ \hline
Task A - OFF & 0.63 & 0.67 & 0.65 \\
Task A - NOT & 0.84 & 0.81 & 0.82 \\
Task B - UNT & 0.21 & 0.48 & 0.30 \\
Task B - TIN & 0.93 & 0.79 & 0.85 \\
Task C - IND & 0.84 & 0.72 & 0.78 \\
Task C - OTH & 0.20  & 0.32 & 0.24 \\
Task C - GRP & 0.54  & 0.61 & 0.57 \\
\hline
\end{tabular}
\end{center}
\caption{Final model validation metrics}
\label{table:final_validation}
\end{table}

\begin{table}[H]
\begin{center}
\begin{tabular}{|l|cc|}
\hline \bf Task & \bf Validation & \bf Official Test \\ \hline
Task A & 0.74 & 0.76 \\
Task B & 0.57 & 0.68 \\
Task C & 0.53 & 0.54 \\
\hline
\end{tabular}
\end{center}
\caption{Final model macro-avg F1 performance}
\label{table:final_test}
\end{table}

Our model performed better in the official challenges than it did in the validation tests. This might be due to the fact that our validation test is relatively small and so doesn't provide a good generalization for the performance of the model.

\section{Discussion}

\subsection{Difficulties encountered}
One of the main difficulties of this task comes from the imbalance in the datasets. We have tried to compensate for this by oversampling the minority classes, undersampling the majority ones, and penalizing the loss of the majority classes. While these solutions did bring some improvements in our results, it remains very complicated to build a model that effectively classifies such imbalanced samples. 

Moreover, we have noticed that some of the tweets in the given training dataset were not correctly labeled (some insults were labelled as NOT and some clearly targeted insults labelled as UNT). For obvious reasons it is not so trivial to estimate the magnitude of this issue and understand how common these mislabelled samples are but, even if quite sparse, they still added onto the overall complexity of the task.

\subsection{Possible improvements}
While the results obtained are very positive, we do realize there is still plenty of room for improvement.
One way to improve the model would be to add more tweets to the FastText. Currently, FastText is applied to the whole corpus of about 13k tweets. Web-scraping for some additional tweets to feed into the algorithm could potentially make the embeddings more accurate and improve the overall performance of the model.

In our study, the proper ratio of under / over-sampling is determined by cross validation using random forest (for time efficiency reason). It would actually be better to do it with the final neural network. Indeed, we have assumed that the shared parameters found optimizing the random forest model will also improve the performance of the final architecture, which might be correct as a general rule, but not necessarily true when trying to find the optimal parameters.

Moreover, the final metric used to evaluate our model in the competition is the macro-average F1. However, the loss function of the neural network is binary cross-entropy. As a result, the neural network tries to maximize the accuracy instead of the F1 score. We have actually tried to implement a loss function close to the average F1-score, but we didn't obtain the desired results. The F1-score function cannot be used by itself as a loss function because it is not differentiable. So, our idea was to compute a loss function close enough to the F1-score: replacing labels 1 and 0 by their softmax probability. However, this approximate loss function was not very effective during the training process and led to worse results than binary cross-entropy. Hence a possible improvement for the future, given some more time to work on the project, would be to design a loss function that effectively maximizes the F1 score

Finally, throughout the project, we focused on the most common techniques used in NLP, but more tailored architectures may work better for this specific task.

\bibliography{semeval2018}
\bibliographystyle{acl_natbib}

\end{document}